\documentclass[sigconf,nonacm]{acmart}
\AtBeginDocument{%
  }

\copyrightyear{2026}
\acmYear{2026}
\setcopyright{cc}
\setcctype{by}
\acmConference[KDD '26]{Proceedings of the 32nd ACM SIGKDD Conference on Knowledge Discovery and Data Mining V.2}{August 09--13, 2026}{Jeju Island, Republic of Korea}
\acmBooktitle{Proceedings of the 32nd ACM SIGKDD Conference on Knowledge Discovery and Data Mining V.2 (KDD '26), August 09--13, 2026, Jeju Island, Republic of Korea}
\acmDOI{10.1145/3770855.3818093}
\acmISBN{979-8-4007-2259-2/2026/08}

\usepackage[utf8]{inputenc}
\usepackage{amsfonts}
\usepackage{amsthm}
\usepackage{mathtools}
\usepackage{makecell}
\usepackage{multirow}
\newcommand{\msd}[2]{\makecell[r]{$#1$\\[-0.2em]{\scriptsize$\pm #2$}}}

\newcommand{\quotes}[1]{``#1''}

\newcommand{\dcs}{\textsuperscript{\scriptsize S}}
\newcommand{\dcm}{\textsuperscript{\scriptsize M}}
\newcommand{\dcl}{\textsuperscript{\scriptsize \textbf{L}}}

\begin{document}

\title{Can Global XAI Methods Reveal Injected Behaviours in LLMs? SHAP vs Rule Extraction vs RuleSHAP}

\author{Francesco Sovrano}
\orcid{0000-0002-6285-1041}

\affiliation{%
  \institution{Collegium Helveticum at ETH Zurich}
  \city{Zurich}
  \country{Switzerland}
}

\affiliation{%
  \institution{Università della Svizzera italiana}
  \city{Lugano}
  \country{Switzerland}
}

\email{francesco.sovrano@usi.ch}

\begin{abstract}
Large language models (LLMs) can amplify misinformation, undermining societal goals such as the UN SDGs.
We study three documented drivers of misinformation (\textit{valence framing}, \textit{information overload}, and \textit{oversimplification}) often shaped by default beliefs.
Building on evidence that LLMs encode such defaults (e.g., \quotes{joy is positive}, \quotes{math is complex}) and can act as \quotes{bags of heuristics}, we ask whether belief-driven heuristics behind misinformation-related behaviour can be recovered from black-box LLM behaviour as explicit rules.
A key obstacle is that global rule-extraction methods in explainable AI (XAI) are built for numerical input-output data, not text. We address this by eliciting global LLM beliefs and mapping them to numerical scores via statistically validated abstractions, enabling off-the-shelf global XAI to detect belief-driven heuristics.
For ground truth, we inject nonlinear behavioural triggers of increasing complexity (univariate, conjunctive, non-convex) into GPT-family and Llama models via system instructions.
We find that \textit{RuleFit} often misses non-univariate triggers, while \textit{global SHAP} better ranks conjunctive trigger features but yields no symbolic rules.
To bridge this gap, we propose \textsc{RuleSHAP}, a rule-extraction algorithm that couples global SHAP aggregates with rule induction to better capture non-univariate triggers, improving MRR@1 over RuleFit by +82\% on average.
Our results suggest a practical pathway for surfacing behavioural triggers in LLMs.
\end{abstract}



\maketitle

\newcommand\kddavailabilitydoi{https://doi.org/10.5281/zenodo.20573706} 
\newcommand\kddrepositoryurl{https://github.com/Francesco-Sovrano/RuleSHAP}
\ifdefempty{\kddavailabilitydoi}{}{%
\begingroup
\small\noindent\raggedright\textbf{Resource Availability:} Source code and artifacts: \url{\kddavailabilitydoi}; repository: \url{\kddrepositoryurl}.
\endgroup
}

\section{Introduction} \label{sec:introduction} 

\begin{figure*}
  \centering
  \includegraphics[width=.65\linewidth]{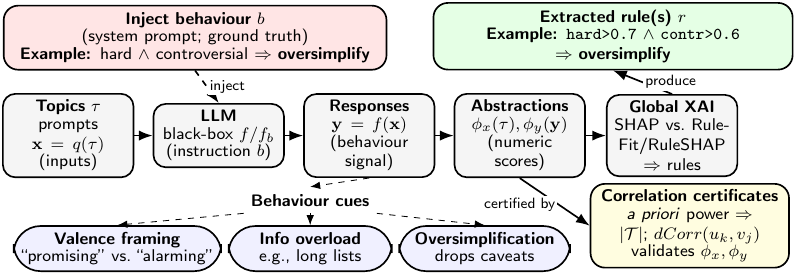}
  \Description{Pipeline diagram showing a controlled LLM behavioural-trigger injection workflow: topics are converted into prompts, rule-gated system instructions induce ground-truth pairs, textual inputs and outputs are converted into numeric abstractions, statistical validation checks the abstractions, and global XAI methods recover auditable rules.}
  \caption{Problem overview: we inject a behavioural trigger $b$ (ground truth) into an LLM, probe with SDG prompts, abstract text into input and output scores, validate abstractions statistically, and apply global XAI to extract human-auditable rules $r$.}
  \label{fig:problem_overview}
\end{figure*}

The \citeauthor{un_sdg_goals}' Sustainable Development Goals (SDGs) \cite{un_sdg_goals} address challenges such as poverty (SDG 1), clean water (SDG 6), and climate action (SDG 13). 
AI systems, and large language models (LLMs) in particular, are increasingly used for knowledge access, with the potential to accelerate or hinder progress towards these societal goals \citep{vinuesa2020role}.
Empirically, their reliability in high-stakes domains remains uneven; e.g., \citet{DBLP:conf/icml/BulianSALCGHBML24} document limitations in generative AI for climate communication. Beyond factual errors, LLMs may also propagate biased or misleading narratives that influence public understanding of SDG-relevant topics \citep{wang2024cognitive}.

We focus on the latter and study three misinformation-related response \emph{behaviours}: \emph{valence framing}, \emph{information overload}, and \emph{oversimplification} \citep{zollo2018misinformation,taslitz2012information}. To avoid overloading the term \emph{bias} (often used for fairness issues), we refer to these mechanisms as \emph{behaviours} and, in controlled experiments, we \emph{inject behavioural triggers} via rule-gated system instructions at inference time (no weight updates) for auditing (Fig.~\ref{fig:problem_overview}).

We cast LLM behavioural auditing as a knowledge discovery problem: mining compact, symbolic rules that describe systematic, prompt-conditioned response patterns.
These behaviours often stem from \textit{default} or \textit{global} beliefs \citep{de2018expectations,han2022default}. For instance, default polarities (e.g., \quotes{disease is negative}) can induce valence framing; perceived complexity (e.g., \quotes{quantum theory is hard}) can trigger overload or oversimplification.
Notably, LLMs have been shown to encode global beliefs \citep{scherrer2023evaluating} and to operate via \quotes{bags of heuristics} \citep{nikankin2024arithmetic}. This motivates our question: can belief-driven heuristics behind these behaviours (e.g., \quotes{if a topic is controversial, oversimplify it}) be recovered from LLMs as symbolic rules?

As shown in Fig.~\ref{fig:problem_overview}, our goal is to reverse-engineer behavioural triggers from observed responses, not to identify weight-level mechanisms.
Global explainable AI (XAI) methods aim to extract such rule sets and are natural candidates, but most are optimized for numerical or tabular data and struggle with text \citep{shruti2024responsible}. For instance, SHAP \citep{lundberg2017unified} uses game theory to identify influential input features yet faces combinatorial explosions over token permutations in vocabulary-sized spaces \citep{bilodeau2024impossibility}. RuleFit \citep{friedman2008predictive}, instead, builds rules via gradient boosting and LASSO pruning, but lacks SHAP's theoretical guarantees and still falters on the input/output spaces of LLMs.

Inspired by evidence that arithmetic prompts can trigger abstraction heuristics in LLMs (e.g., parity or range cues \citep{nikankin2024arithmetic}), we introduce a statistically grounded \emph{abstraction} pipeline. It maps LLM-expressed topic beliefs and generated explanations to input and output abstractions, enabling global XAI tools to detect belief-driven heuristics that shape behaviour. To produce explanations that capture stable input-output regularities, we collect LLM inputs and outputs, with sample sizes set by \emph{a priori} power analysis. We then use \emph{correlation certificates} to test whether the extracted rules and abstractions align with statistically detectable behavioural regularities.

We evaluate state-of-the-art global XAI methods (with publicly available implementations) on popular LLM families (ChatGPT and Llama) and on SDG-related topics. 
To establish ground truth, we inject 14 behavioural triggers at inference time using three conditional system instruction templates (no weight updates).
To evaluate performance under varying nonlinearities, we analyse trigger predicates of increasing complexity: univariate, conjunctive, and non-convex. We then assess how many ground-truth triggers are recovered by the XAI methods.
%
While global SHAP flags injected behavioural triggers, it cannot express it as rules. RuleFit instead overproduces rules and misses many non-univariate effects. We thus introduce \textsc{RuleSHAP}, which refines RuleFit's rule generation via global SHAP aggregates. In controlled injections, \textsc{RuleSHAP} improves average MRR@1 from 0.314 for RuleFit to 0.571, and reduces the number of extracted rules by 12.4\% relative to RuleFit. The improvement over RuleFit is significant for RR@1, RR@3, and RR@10 (one-sided Wilcoxon signed-rank tests, Holm--Bonferroni corrected; all adjusted $p<0.005$).  
Notably, rule-detection accuracy falls as rule complexity increases, and even our injected behavioural triggers (simpler than many real-world cases) challenge state-of-the-art XAI, exposing critical limitations for future work to address.
Finally, we show that recovered rules can be used as conditional triggers for lightweight steering in a no-injection case study.

\textbf{Contributions:}
We formulate LLM behavioural auditing as global rule recovery over ordered input and output abstractions; 
introduce a controlled benchmark with 14 inference-time behavioural triggers spanning univariate, conjunctive, and non-convex forms; 
propose \textsc{RuleSHAP}, which couples global SHAP aggregates with rule induction and sparse rule selection; and 
evaluate exact recovery, functional fidelity, abstraction stability, imputation robustness, no-injection generalization, and rule-gated steering across LLMs.

\section{Related Work} \label{sec:related_work}

Global, model-agnostic XAI methods that recover nonlinear \emph{rules} for generic input-output data remain uncommon (\S\ref{sec:experiments:baselines}); most rely on tree surrogates. SHAP-based global explainers \citep{lundberg2019explainable,watson2022rational,zhao2024shapg} provide importance attributions but not symbolic rules, while RuleFit-style methods provide rules without SHAP's feature-credit guarantees. \textsc{RuleSHAP} is designed for the intersection: it uses global SHAP aggregates to steer both rule generation and sparse rule selection.

Recent NLP-specific global explainers are closest in motivation. GELPE builds small logic programs for text classification by composing local post-hoc explanations into a CART-style symbolic surrogate \citep{agiollo2024gelpe}. GloVE extracts verifiable global policies for LLM-as-a-judge systems by clustering, summarizing, and verifying local concept explanations \citep{gajcin2025glove}. Our setting differs in three ways: we audit black-box LLM \emph{behavioural triggers} rather than text-classification or judge policies; inputs and outputs are graded abstraction scores rather than binary lexical indicators or local concept rules; and the controlled benchmark varies trigger complexity (univariate, conjunctive, non-convex) so that exact recovery, functional fidelity, and degradation under harder rules can be measured.

For autoregressive generation, token-level SHAP faces combinatorial explosions over vocabularies \citep{bilodeau2024impossibility,goldshmidt2024tokenshap}. Concept-level or abstraction-level alternatives can reduce this burden \citep{amara2025concept}; our pipeline adopts this idea globally by mapping topics and outputs to ordered abstractions before applying XAI. Mechanistic interpretability instead seeks neural circuits, but such circuits do not always yield symbolic global behavioural rules \citep{dunefsky2024transcoders}; for arithmetic, LLMs have nevertheless been shown to rely on identifiable heuristic-like features \citep{nikankin2024arithmetic}.
Finally, most LLM-bias studies use mixed-methods and centre on human--LLM alignment rather than extracting and ranking behavioural triggers; e.g., \citet{DBLP:conf/icml/BulianSALCGHBML24} study climate information, \citet{wu2024surveying} survey SDG alignment, and \citet{kumar2024decoding,koo2023benchmarking} study LLMs as evaluators. Our SDG framing provides a diverse social testbed; the technical contribution is the abstraction-plus-rule-recovery framework.

\section{Belief Abstraction Pipeline for Rule Extraction in LLMs} \label{sec:tom} \label{sec:tom:preliminaries} \label{sec:tom:problem_formulation} \label{sec:tom:framework}

\textbf{Preliminaries: Global XAI.}
Two model-agnostic XAI paradigms underpin our work: \emph{perturbation-based} techniques and \emph{rule-extraction} methods.
Perturbation-based methods, e.g., LIME \citep{ribeiro2016model} and SHAP \citep{lundberg2017unified}, analyse model responses to perturbed inputs \( x' \) generated by modifying subsets of features in \( x \!\in\! \mathbb{R}^d \). SHAP estimates the feature importance \( \rho_i \) to the output, satisfying \textit{local accuracy} (additivity): \( f(x) \!=\! \sum_{i=0}^d \rho_i \). 
Rule-extraction methods, such as RuleFit \citep{friedman2008predictive}, instead derive interpretable rules \( R \!=\! \{r_1, \ldots, r_k\} \) from a dataset \( X \) using surrogate models \( f_s \) (e.g., decision trees), approximating the original model \( f \) via \( f_s(X) \!\approx\! f(X) \).
Both approaches generally assume input features suitable for perturbation and outputs with an inherent ordering, e.g., logits, log-odds, or probabilities.

\textbf{Data selection.}
In this paper, we use global XAI to surface high-level LLM heuristics that can propagate misinformation \citep{zollo2018misinformation} by triggering three cognitive behaviours: \textit{framing effects} (selective valence framing of attributes), \textit{information overload} (overly long or complex outputs), and \textit{oversimplification} (superficial explanations of complex issues).
Existing datasets on cognitive bias in LLMs are confined to medicine \citep{schmidgall2024addressing, wang2024cognitive} and software engineering \citep{sovrano2025general,sovrano2026mitigating}, and, critically, do not expose the underlying behavioural triggers themselves. This lack of ground-truth rules prevents validation of explanations produced by global XAI; without them, the faithfulness of such explanations cannot be assessed.

Since no suitable dataset exists for our experiments, we focus on detecting behavioural triggers that shape LLM-generated explanations of SDG-related topics, due to their global impact (e.g., misinformation on poverty or gender equality can have UN-wide implications) and the ease of generating related prompts.
Formally, let $\mathcal{T} = \{\tau_1, \dots, \tau_{|\mathcal{T}|}\}$ be a set of \emph{topics} (i.e., textual phrases), each tied to an SDG-related challenge or solution (e.g., ``micro-algae for CO\textsubscript{2} absorption'').
Let $\mathcal{X} = \{\mathbf{x}_1, \dots, \mathbf{x}_N\} \subset \Sigma^*$ denote a set of prompt texts, each being a request for an explanation of a topic $\tau_i \in \mathcal{T}$ generated by a function $q \colon \mathcal{T} \to \mathcal{X}$ such that $\mathbf{x}_i = q(\tau_i)$, where $q(\tau_i) =$ \quotes{\textit{Explain \{$\tau_i$\}}} is the simplest choice.
Then, $f : \mathcal{X} \to \Sigma^*$ models an LLM producing a textual explanation $\mathbf{y}_i = f(q(\tau_i))$ for each topic $\tau_i$. 

\textbf{Abstraction setting.}
Motivated by evidence that LLMs encode data-induced global beliefs that can align with human judgments \citep{lee2025semantic,scherrer2023evaluating}, we extract input abstractions from \emph{topics} (e.g., \quotes{$\tau_i$ is common}) and output abstractions from LLM-generated \emph{explanations} (e.g., \quotes{$\mathbf{y}_i$ is unreadable}).
Specifically, we define two mappings, \(\phi_x \colon \mathcal{T} \!\to\! \mathbb{R}^d\) and \(\phi_y \colon \Sigma^* \!\to\! \mathbb{R}^m\), that transform textual data into numerical (ordered) representations: $\mathbf{u}_i \!=\! \phi_x(\tau_i) \in \mathbb{R}^d$ and $\mathbf{v}_i \!=\! \phi_y(\mathbf{y}_i) \in \mathbb{R}^m$.
Once these mappings are applied, the textual prompts and responses (\(\mathbf{x}_i\), \(\mathbf{y}_i\)) are no longer used in the subsequent XAI pipeline. Thereby, perturbation-based or rule-extraction XAI methods operate directly on the numeric vectors \(\{\mathbf{u}_i, \mathbf{v}_i\}\). 

\textbf{Main assumption and important workaround.} 
The framework above assumes that topics with identical \(\mathbf{u}\)-vectors (\(\mathbf{u}_i = \mathbf{u}_j\)) yield similar output vectors \(\mathbf{v}_i\) and \(\mathbf{v}_j\), an assumption that can fail because textual generation and scoring are not perfectly stable.
To address this, a sufficiently large \(\mathcal{T}\) is needed to ensure statistical power and \textit{detectable correlations} between \(\mathbf{u}_i\) and \(\mathbf{v}_i\). These correlations are estimated via Pearson or Spearman methods for linear or monotonic regularities, or via \textit{distance correlation} \citep{szekely2009brownian} for nonlinear regularities (as in our case). 
Thus, \emph{global XAI insights from text-to-ordered mappings should be supported by correlation evidence showing that the abstractions capture the intended regularities}.

\textbf{Input abstractions.}
We analyse SDG-related topics $\tau_i \in \mathcal{T}$. Each topic $\tau_i$ is mapped to an ordered feature vector $\mathbf{u}_i = \phi_x(\tau_i) \in \mathbb{R}^{d=11}$. The vector $\mathbf{u}_i$ represents the audited LLM's expressed belief about $\tau_i$ under a fixed scoring prompt; operationally, $\phi_x(\cdot)$ is implemented by querying that same LLM with deterministic decoding ($T=0$, top-$p=0$). For each SDG topic, the prompt asks the model to estimate how web text typically discusses the topic. A raw score $s\in\{1,2,3,4,5\}$ is stored as $s/5$, so each topic dimension lies in $\{0.2,0.4,0.6,0.8,1.0\}$.
The 11 dimensions are: \emph{conceptual density} (layered or abstract ideas), \emph{technical complication} (specialist terminology or procedures), \emph{commonality} (prevalence and general familiarity), \emph{social controversy} (polarization), \emph{unambiguity} (stable interpretation across contexts), \emph{geographic variability} (changes across regions), \emph{time variability} (changes over time), \emph{interdisciplinarity} (span across fields), and \emph{positivity}, \emph{negativity}, and \emph{neutrality} (valence in online discourse). Thus, for each topic $\tau_i$, $\mathbf{u}_i=\phi_x(\tau_i)$ is an 11-dimensional ordered vector. Appendix~\ref{apx:abstractions} gives the operational prompt suffixes and score handling details.

\textbf{Output abstractions.}
We instantiate the output abstraction function as $\phi_y:\Sigma^*\to\mathbb R^{m=7}$. It maps each generated explanation $\mathbf y_i=f(q(\tau_i))$ to four non-judge proxies and three supplementary LLM-as-a-judge scores. The non-judge proxies are character length, which reflects possible information overload when high and oversimplification when low \citep{arnold2023dealing,che2019information}; Gunning--Fog readability, where higher values indicate more difficult prose \citep{gunning1969fog}; sentiment polarity from TabularisAI's publicly available multilingual-sentiment Transformer estimator; and subjectivity from GroNLP's publicly available multilingual mDeBERTa-v3 subjectivity estimator. Sentiment and subjectivity serve as proxies for valence framing \citep{capraro2019power}. Long explanations are split into approximately 400-token chunks for Transformer scoring. For sentiment, we retain the maximum confidence for each polarity over chunks and encode the dominant class as a signed score: positive confidence, negative confidence with a minus sign, or 0 when the neutral class dominates. For subjectivity, we use the maximum subjective-class confidence over chunks, yielding a score in $[0,1]$. The supplementary judge scores rate \emph{framing effect}, \emph{information overload}, and \emph{oversimplification} on deterministic 1--5 scales using an LLM-as-a-judge protocol \citep{zheng2023judging,thakur2024judging}. To avoid circularity, judge-derived scores are interpreted only as supplementary signals and if they agree with the non-judge proxies.

\section{RuleSHAP} \label{sec:ruleshap} \label{sec:ruleshap:preliminaries} \label{sec:ruleshap:implementation}

We briefly review SHAP and RuleFit, then describe \textsc{RuleSHAP}.

\textbf{SHAP} \citep{lundberg2017unified} is a model-agnostic, game-theory-based XAI method. Given a model $f$ and instance $u$, it assigns each feature $u_i$ a Shapley value $\rho_i$, representing its average contribution to $f(u)$ across all feature coalitions. This is estimated by replacing features with predefined \textit{background} values (not necessarily $0$) to measure the impact of their removal. For global attributions, one can average SHAP's local attributions \citep{mayer2023shap}, providing an overall feature-importance measure.
Since exact SHAP is computationally expensive and assumes feature independence, applications often rely on approximations such as KernelSHAP (weighted linear regression) or the permutation SHAP explainer (used in our experiments), which approximates Shapley values by sampling feature permutations (potentially underrepresenting higher-order interactions). To address bias from violated independence assumptions, the official SHAP documentation \citep{shap_partition_explainer} recommends grouping correlated features through hierarchical partitioning (as we do). These clusters are defined from feature dependence conditioned on the target, with joint credit assigned at the cluster level.

\textbf{RuleFit} \citep{friedman2008predictive} is a global XAI method that extracts the rules a model \( f \) follows to produce outputs \(\mathbf{v}\) from inputs \(\mathbf{u}\). It does this by leveraging features within \(\mathbf{u}\), where each input vector consists of multiple features \( c_1, c_2, \dots, c_d \).
RuleFit combines gradient boosting \citep{natekin2013gradient} with sparse linear modelling to build interpretable models. Gradient boosting iteratively constructs decision trees, where each split minimizes impurity or maximizes loss reduction. A decision path from root to leaf defines a binary rule, e.g., $(c_i \leq \theta_1) \land (c_j > \theta_2) \land \dots$, where each $\theta$ is a threshold value.
After extracting rules, RuleFit forms a binary design matrix \(\mathbf{X}\in\{0,1\}^{N\times |R|}\), with rows indexing instances and columns indexing rules. Each entry \(X_{n,k}\) is 1 if rule \(r_k\) fires on instance \(u_n\), otherwise 0. Thus, \(\mathbf{X}\) captures the activation of each extracted rule across the dataset.
Then, given the coefficient vector \(\mathbf{w}\) and a hyperparameter \(\alpha\) to control for sparsity, RuleFit applies a LASSO regression \citep{ranstam2018lasso} to select a sparse set of predictive rules by solving:
$
\min_{\mathbf{w}} \ \frac{1}{2}\|\mathbf{v} - \mathbf{X} \mathbf{w}\|_2^2 + \alpha \|\mathbf{w}\|_1.
$

\textbf{Technical Challenges and Proposed Solution.}
SHAP provides rigorous game-theoretic feature attributions with additivity and consistency but lacks global symbolic representations, limiting its utility when human-readable rules are needed, particularly for complex relationships. RuleFit generates interpretable rules with gradient-boosted trees and LASSO but lacks SHAP's theoretical guarantees, so its rules may not reflect a truly global measure of feature importance.
To overcome the limitations of SHAP and RuleFit, we introduce \textsc{RuleSHAP}, a novel algorithm integrating SHAP into RuleFit. This integration leverages SHAP's robust feature attribution, preserving RuleFit's interpretability and rule-based representation. To achieve this, we follow these steps:

\textbf{Step 1: Global Shapley Values Aggregation.}
We first compute SHAP values for all available data points with a standard SHAP explainer.
Specifically, for each feature \( c_i \) and instance \( \mathbf{u}_k \), we obtain the Shapley value \( \rho_{(i,k)} \), using a near-minimum feature vector as the \textit{background}: each coordinate is set slightly below the observed minimum for that feature. This baseline is meaningful because each input abstraction is an ordinal degree-of-presence score, so the near-minimum reference represents the least-expressed in-distribution state for that property rather than an arbitrary zero.
Because our features are abstractions of text I/O, we cannot implement SHAP's usual feature masking by constructing counterfactual prompts and re-querying the LLM. 
%
Instead, following the workaround from \S\ref{sec:tom:preliminaries}, we ensure sufficient redundancy in \( \mathcal{T} \) so that, for each SHAP perturbation \( \widehat{\mathbf{u}}_k \), we can find one or more nearest neighbours \( \mathbf{u}_j \!\in\! \mathcal{T} \) (ideally, with \( \|\widehat{\mathbf{u}}_k - \mathbf{u}_j\|_2 \!\approx\! 0 \)) and matching the background values used by SHAP. We then impute each abstract perturbation with one of these neighbours, allowing contribution estimation without additional LLM calls while staying within the observed abstraction distribution.
%
Specifically, we aggregate the \( \rho_{(i,k)} \) values across all instances by calculating the mean of the absolute Shapley values, adding the standard deviation to capture the upper bound of each feature's importance:
$
\rho_i^{\text{agg}} \!=\! \frac{1}{N} \sum_{k=1}^{N} |\rho_{(i,k)}| + \text{std}( \{|\rho_{(i,k)}|\}_{k=1}^N )
$,
where \( N \) is the number of instances. 
The aggregated Shapley values \( \rho_i^{\text{agg}} \) are then normalized by $\bar{\rho}_i \!=\! \rho_i^{\text{agg}} / \sum_{j=1}^{d} \rho_j^{\text{agg}}$, where \( d \) is the number of input features.

\textbf{Step 2: XGBoost with SHAP Weighting.}
We replace RuleFit's gradient-boosting component with XGBoost \citep{chen2016xgboost} and use the aggregated Shapley values from Step~1 as feature-sampling priorities. Thus, each feature \(c_i\) is sampled with priority proportional to its global SHAP weight \(\bar{\rho}_i\).
To integrate SHAP-driven rule selection into XGBoost, we have to adjust specific hyperparameters to leverage feature weights effectively. First, we configure \texttt{colsample\_bylevel} to \( \frac{1}{d} \), which controls the fraction of features sampled at each tree level. By setting this parameter accordingly, the model is encouraged to sample approximately one feature per level. Specifically, this sampling process is guided by the computed feature weights \( \{ \bar{\rho}_i \} \), meaning that features with higher Shapley values have a greater likelihood of being selected at each level. This prioritization encourages globally important features to play a larger role in model construction. 
Additionally, we employ the \textit{exact} setting for the \texttt{tree\_method} hyperparameter, which directs XGBoost to use a precise greedy algorithm that thoroughly evaluates all possible split points across selected features. This setting evaluates exact split gains within the selected feature subset at each node, reducing the risk of suboptimal partitioning that could otherwise arise from the low \texttt{colsample\_bylevel} value.

\textbf{Step 3: SHAP-Aware LASSO Regression.} 
Finally, we modify the LASSO regression component to prioritize rules that involve features with higher global aggregated Shapley values. For each extracted rule \( r_j \), identify the set of features \( \mathcal{F}(r_j) \) that appear in its conditions. Compute the average aggregated Shapley value for these features: $\rho_{r_j} = \frac{1}{|\mathcal{F}(r_j)|} \sum_{c_i \in \mathcal{F}(r_j)} \bar{\rho}_i$.
These rule-specific weights \( \rho_{r_j} \) reflect the overall importance of the features involved in each rule, prioritizing rules whose constituent features have high average global SHAP mass.
We then adjust the LASSO regularization term to incorporate these weights, modifying the optimization problem as follows:
$
\min_{\mathbf{w}} \ \frac{1}{2} \|\mathbf{v} - \mathbf{X} \mathbf{w}\|_2^2 + \alpha \sum_{j=1}^{|R|} \frac{|\mathbf{w}_j|}{\rho_{r_j}+\epsilon}.
$
By scaling the regularization term inversely with \( \rho_{r_j}+\epsilon \), where \(\epsilon>0\) prevents division by zero, rules associated with more important features are less penalized, thereby being more likely to be retained in the final model. This approach encourages selected rules to capture significant feature interactions while aligning rule selection with the global importance measures provided by SHAP.

\textbf{Step 4: Rule Importance Computation.}
Additionally, unlike RuleFit, we compute the importance of a rule \( r_j \) using only the absolute value of its coefficient learned via LASSO regression, i.e., $I(r_j) = |w_j|$, where \( w_j \) is the coefficient assigned to rule \( r_j \) by the LASSO model. In contrast, RuleFit defines rule importance as
$
I_{\text{RuleFit}}(r_j) = |w_j| \cdot \sqrt{\text{support}(r_j) \cdot (1 - \text{support}(r_j))}
$,
where \( \text{support}(r_j) \) represents the proportion of training samples that satisfy rule \( r_j \). We avoid this support-based weighting because it inherently penalizes rules with either very high or low support. This can be problematic, as some injected behavioural triggers may affect only a small subset of instances. 

\textbf{Algorithmic summary.}
Given paired abstractions $\{(\mathbf{u}_i,\mathbf{v}_i)\}_{i=1}^N$ and a target output abstraction $v_j$, \textsc{RuleSHAP}: \textit{(1)} fits the target model and computes local SHAP values over the observed abstraction space; \textit{(2)} aggregates them into normalized global feature weights $\bar{\rho}_i$; \textit{(3)} trains XGBoost with feature sampling biased by $\bar{\rho}_i$ and extracts decision-path rules; \textit{(4)} builds the binary rule-activation matrix $\mathbf{X}$; \textit{(5)} solves the SHAP-aware weighted LASSO; and \textit{(6)} ranks retained rules by $|w_j|$. Thus, SHAP enters twice: before tree construction, to increase exposure of globally relevant features, and during sparse selection, to penalize rules built on low-importance features.

\textbf{Theoretical justifications.}
Appendix~\ref{apx:ruleshap-proofs} gives compact proof sketches showing that the two SHAP-weighted steps are beneficial rather than unconditional guarantees: feature sampling improves relevant-feature coverage when SHAP mass is concentrated on relevant dimensions, while the sparse-selection stage matches an adaptive-LASSO form under standard sparsity, design, and tuning assumptions \citep{zou2006adaptive}.

\section{Experimental Setup} \label{sec:experiments} \label{sec:experiments:topics} \label{sec:experiments:biases} \label{sec:experiments:baselines}

We evaluate the effectiveness of global XAI in detecting belief-driven heuristics across five popular general-purpose LLMs (GPT-4o, GPT-4o mini, GPT-3.5 Turbo, Llama-3.1~8B, Llama-3.1~70B).
Fig.~\ref{fig:problem_overview} summarizes our controlled auditing setup: we first sample a (large) set of (SDG-related) topics $\mathcal{T}$ (\S\ref{sec:tom}). To obtain ground truth, we apply conditional system-instruction templates $b_j$ to a base LLM $f$, producing behaviour-modified models $f_{b_j}$. The trigger predicates induced by these templates vary in complexity, incorporating nonlinear relationships of three different types: \textit{univariate}, \textit{conjunctive} multivariate, and \textit{non-convex}/disconnected (e.g., modulus operator).
Using $b_j$, we generate explanations $\mathbf{y}_i \!=\! f_{b_j}(\mathbf{x}_i)$ and compute input and output abstractions via $\phi_x(\tau_i)$ and $\phi_y(\mathbf{y}_i)$. 
Finally, we apply global XAI (e.g., \textsc{RuleSHAP}) to each behaviour-modified model $f_{b_j}$ to detect injected behavioural triggers by analysing LLM behaviour \textit{across all topics $\mathcal{T}$}. Further experimental setup details follow.

    
    
    

\textbf{Topic extraction.}
For the topic selection phase, we focus on three domains, corresponding to distinct UN SDGs: \emph{climate action} (SDG 13), \emph{good health and well-being} (SDG 3), and \emph{gender equality} (SDG 5). These domains were chosen for their global relevance and the diverse range of topics they encompass.
Within each domain, we categorize topics according to the dimensions defined in \S\ref{sec:tom:framework}, i.e., whether a topic is \emph{conceptually dense}, \emph{technically complicated}, etc. 

Following the workaround in \S\ref{sec:tom}, we set the sampling target for $\mathcal{T}$ through an \textit{a priori} power analysis for detecting a medium bivariate association, with target power $1 \!-\! \beta \!=\! 0.8$, effect size $\rho \!=\! 0.35$, two-tailed $\alpha \!=\! 0.05$, and null correlation $\rho_0 \!=\! 0$. This yields a required sample size of $N \!=\! 61$ for the planned correlation tests.
%
We therefore generated, with each LLM, at least 60 topics per domain, score level, and seed descriptor. The generation descriptors included the 11 audited abstraction dimensions plus auxiliary coverage descriptors used only to broaden topic coverage; after de-duplication, all retained topics were re-scored on the same 11 audited dimensions.

Topic generation used $T \!=\! 1$ and top-$p \!=\! 1$ to encourage diversity, while abstraction scoring, judge scoring, injected-response generation, and targeted steering generations used deterministic decoding ($T \!=\! 0$, top-$p \!=\! 0$). Because each topic is scored across all 11 abstraction dimensions, stratified sampling and subsequent de-duplication yield a topic pool with broad cross-dimensional coverage. The final number of unique topics per model is reported in Table~\ref{tab:unique_topics} and exceeds the minimum required by the power analysis.
In other words, if many similar data points (in the abstracted input space) yield outputs with the same output abstractions, it becomes increasingly likely that their abstractions are truly correlated with these properties.

Near-duplicate topics were removed by calculating semantic similarity among all topics with \texttt{all-MiniLM-L6-v2} \citep{reimers-2019-sentence-bert} and a cosine-similarity threshold of 90\%. A sensitivity check with two encoders (\texttt{all-MiniLM-L6-v2}, \texttt{all-mpnet-base-v2}) and thresholds 0.85, 0.90, and 0.95 produced mean retention 0.921 and mean pairwise Jaccard overlap 0.874, indicating that the retained topic pool is not driven by a single de-duplication setting. Appendix~\ref{apx:hyperparameters} collects the main decoding, sampling, and rule-extraction hyperparameters.

\begin{table}[tbh]
    \centering
    \resizebox{.88\linewidth}{!}
    {%
    	\begin{tabular}{lcrrrrr} 
    		\toprule
    		SDG            & \multicolumn{1}{l}{ID} & \multicolumn{1}{c}{\begin{tabular}[c]{@{}c@{}}GPT-3.5~\\Turbo\end{tabular}} & \multicolumn{1}{c}{\begin{tabular}[c]{@{}c@{}}GPT-4o~\\mini\end{tabular}} & \multicolumn{1}{c}{GPT-4o} & \multicolumn{1}{c}{\begin{tabular}[c]{@{}c@{}}Llama\\3.1~8B\end{tabular}} & \multicolumn{1}{c}{\begin{tabular}[c]{@{}c@{}}Llama\\3.1~70B\end{tabular}}  \\ 
    		\midrule
    		Climate Act.   & 13                     & 1,372                             & 2,275                           & 2,870                      & 2,974                         & 1,864                              \\
    		Well-Being     & 3                      & 1,510                             & 2,488                           & 3,391                      & 3,903                         & 3,382                              \\
    		Gender Eq.     & 5                      & 1,297                             & 2,095                           & 2,585                      & 3,667                         & 2,932                              \\ 
    		\midrule
    		\textbf{Total} &                        & \textbf{4,179}                    & \textbf{6,858}                  & \textbf{8,846}             & \textbf{10,544}               & \textbf{8,178}                     \\
    		\bottomrule
    	\end{tabular}
    }
    \caption{Unique topics found per LLM and SDG domain.} \label{tab:unique_topics}
\end{table}

\textbf{Injecting behaviours.}
Let $B=\{b_1,b_2,b_3\}$ be three conditional system-instruction templates with increasingly nonlinear trigger predicates. The behaviour-modified LLM under template $b_j$ is denoted $f_{b_j}$. Input abstractions are computed once with the base scoring prompt and reused across $b_1$--$b_3$, rather than recomputed under injected instructions. At inference time, the wrapper appends the template's instruction iff the precomputed abstraction $\mathbf{u}_i=\phi_x(\tau_i)$ satisfies its trigger predicate, where $\mathbf{x}_i=q(\tau_i)$. Each template defines one or more ground-truth pairs, each consisting of a trigger predicate and an affected output abstraction. Across $b_1$--$b_3$, we obtain 14 pairs: 3 univariate, 5 conjunctive, and 6 non-convex. In rule matching, thresholds are expressed at midpoints between adjacent Likert values (e.g., $p_{\mathrm{raw}}\ge3$ as $p>0.5$), matching the form emitted by tree-based extractors. We denote ground-truth pairs by $g\in G$ and reserve $r$ for XAI-extracted rules.

The simplest trigger template, \( b_1 \), adjusts response length based on the topic's raw commonality score \( c_{\mathrm{raw}}(\tau_i)\in\{1,\ldots,5\} \), or equivalently on its normalized score \(c(\tau_i)=c_{\mathrm{raw}}(\tau_i)/5\). Topics with lower commonality (\( c_{\mathrm{raw}}(\tau_i) \leq 4\), equivalently \(c(\tau_i)\leq0.8\)) receive a concise, single-sentence instruction, whereas more common topics receive an extended, multi-paragraph instruction. 
Thus, \(b_1\) defines one univariate trigger predicate over commonality. Because this predicate is a single upper-bound constraint, it is monotonic and convex, in a mathematical sense; evaluated against three target output abstractions, it yields three ground-truth pairs.

Building on this, the \textit{conjunctive} template \(b_2\) is multivariate, i.e., it conditions response style on commonality and positivity. Topics with low commonality and moderate positivity (\(c_{\mathrm{raw}}(\tau_i)\leq2 \land p_{\mathrm{raw}}(\tau_i)\geq3\), equivalently \(c(\tau_i)\leq0.4 \land p(\tau_i)\geq0.6\)) elicit multi-paragraph explanations. If positivity is higher (\(p_{\mathrm{raw}}(\tau_i)\geq4\), equivalently \(p(\tau_i)\geq0.8\)), an additional directive prompts subjective negative framing. Thus, \(b_2\) is convex, as an intersection of half-spaces, and yields five ground-truth pairs: two with non-judge output abstractions and three with judge-derived scores.

The most complex template, \( b_3 \), extends \( b_2 \) with a \textit{non-convex} readability degradation rule conditioned on interdisciplinarity \( d(\tau_i) \).
Concretely, when the raw Likert value of interdisciplinarity is in \(\{1,3,5\}\), equivalently when the normalized score is in \(\{0.2,0.6,1.0\}\), the model is directed to use convoluted, overly complex language to reduce readability, increasing the risk of information overload and oversimplification. 
This introduces disconnected regions in the trigger predicate, making \( b_3 \) non-convex, and yields six ground-truth pairs: three involving non-judge output abstractions and three involving judge-derived scores. For $b_3$, topics whose raw interdisciplinarity score falls in the disconnected set $\{1,3,5\}$ receive more convoluted wording, making readability degradation harder to recover as a single convex rule. Appendix~\ref{apx:concrete_examples} gives topic-level examples.

\textbf{We decode explanations and abstraction scores deterministically} (greedy decoding; temperature $T\!=\!0$), so outputs reflect instructions and model parameters rather than sampling noise. Higher temperatures ($T\!=\!2$) are possible, but they can induce off-instruction drift, especially in smaller models such as Llama-3.1~8B, and therefore weaken correlation certificates. We treat rules with weak certificates as unreliable rather than as validated behavioural explanations (\S\ref{sec:tom} and \S\ref{sec:results}).

\textbf{Baseline XAI methods.}
Global rule extraction methods such as GLocalX \citep{setzu2021glocalx}, Skope-Rules \citep{singh2021imodels}, Bayesian Rule Sets \citep{7837984}, and FIGS \citep{tan2023fast} are widely used for models with categorical outputs.
However, we focus on more generic, numerical outputs, where the landscape of XAI solutions is sparser. In this case, researchers typically rely on linear regression or other tree-based methods (e.g., decision trees).
Although linear regression can identify linear effects, it cannot capture the nonlinear trigger predicates defined in \S\ref{sec:experiments:biases}, so we do not treat it as a rule-extraction baseline.
Decision trees, instead, can handle nonlinearity. Methods like MRE \citep{asano2021post} and PALM \citep{krishnan2017palm} use them but lack implementation code, so are excluded.
NLP-specific methods such as GELPE \citep{agiollo2024gelpe} and GloVE \citep{gajcin2025glove} are discussed as related work rather than treated as primary baselines, because GELPE targets only binary lexical indicators in text classification and GloVE lacks implementation code.

Hence, we use RuleFit \citep{friedman2008predictive} (which relies on gradient boosting and is open source) and decision-tree-based surrogate models as baselines. As an additional baseline, we incorporate RuleFit enhanced with XGBoost instead of traditional gradient boosting, which is equivalent to \textsc{RuleSHAP} with uniform feature weights. Despite SHAP not producing explicit rule-based explanations, we also include it in a separate evaluation to understand to what extent its global explanations can capture the importance of features involved in trigger predicates.
To perform an ablation study on \textsc{RuleSHAP}'s main components, we consider two modified versions of \textsc{RuleSHAP}. The first variant omits Step 2, excluding SHAP-driven feature weighting in XGBoost. The second variant excludes Step 3, i.e., no SHAP-aware LASSO. 

\textbf{Evaluation metrics.}
We report three complementary metrics. 
First, \textit{exact recovery} is measured by mean reciprocal rank (MRR) at $k\in\{1,3,10\}$. For injected triggers $G$ and extracted rules $R$, let $r(g)$ be the rank of the first rule whose predicates and target output abstraction exactly match trigger $g$. We define $\mathrm{RR@}k(g)=\mathbf{1}[r(g)\le k]/r(g)$ and $\mathrm{MRR@}k=|G|^{-1}\sum_{g\in G}\mathrm{RR@}k(g)$. This metric is conservative because it rewards syntactic recovery and can miss functionally equivalent rules.


Second, \textit{functional fidelity} measures whether the highest-ranked extracted rules identify the same instances as the planted trigger, even when the recovered rule is not syntactically identical to the ground truth. For each ground-truth pair and cutoff \(k\), we take the top-\(k\) extracted rules for that target and form a binary surrogate that fires on an instance when at least one of those rules fires. We then compare this surrogate firing set with the ground-truth triggered subset induced by the injected predicate. We report Matthews correlation coefficient (MCC), using it as the primary fidelity metric because it remains informative under class imbalance.

Third, \textit{rule-set size} $|R|$ measures compactness. We interpret it together with exact and functional fidelity, since smaller rule sets are useful only when they preserve behaviour.

\section{Results and Analysis} \label{sec:results}

We first assess how well SHAP highlights trigger-relevant features, then compare \textsc{RuleSHAP} to baselines on interpretability (rule compactness) and behaviour auditing (MRR). 

\textbf{Global SHAP.}
We measure SHAP feature ranking with the same MRR metrics used for exact rule recovery: a feature receives credit when a behaviour-relevant input dimension appears within the top-$k$ ranked dimensions. At MRR@1, SHAP scores 0.42 on each GPT model and 0.33 on each Llama model. At MRR@3, the scores are 0.49 for GPT-3.5 Turbo, 0.60 for GPT-4o mini, 0.53 for GPT-4o, 0.47 for Llama-3.1~8B, and 0.44 for Llama-3.1~70B. At MRR@10, the corresponding scores are 0.55, 0.62, 0.59, 0.55, and 0.51. Sorting feature importance by upper bound (mean + standard deviation) yields the highest MRRs, empirically outperforming mean or max and motivating its use in \textsc{RuleSHAP} (\S\ref{sec:ruleshap:implementation}).

\begin{table}[tbh]
    \centering
    \scriptsize
    \setlength{\tabcolsep}{3pt}
    \resizebox{.99\linewidth}{!}{%
       \begin{tabular}{llrrrr}
        \toprule
        \textbf{LLM} & \textbf{XAI Method} & \multicolumn{1}{c}{\textbf{Rules}} & \multicolumn{1}{c}{\textbf{MRR@1}} & \multicolumn{1}{c}{\textbf{MRR@3}} & \multicolumn{1}{c}{\textbf{MRR@10}} \\
        \midrule
        \multirow{6}{*}{\begin{tabular}[c]{@{}l@{}}GPT-3.5\\Turbo\end{tabular}}
        & Decision Tree & 987 & 0.00 & 0.00 & 0.00 \\
        & RuleFit & 1096 & 0.29 & 0.31 & 0.33 \\
        & RuleFit+XGB & 1429 & 0.14 & 0.14 & 0.14 \\
        & RuleSHAP w/o Step 2 & 923 & 0.14 & 0.14 & 0.14 \\
        & RuleSHAP w/o Step 3 & 1086 & 0.29 & 0.32 & 0.34 \\
        & RuleSHAP & 943 & \textbf{0.36} & \textbf{0.45} & \textbf{0.47} \\
        \midrule
        \multirow{6}{*}{\begin{tabular}[c]{@{}l@{}}GPT-4o\\mini\end{tabular}}
        & Decision Tree & 1005 & 0.00 & 0.00 & 0.00 \\
        & RuleFit & 1681 & 0.36 & 0.38 & 0.40 \\
        & RuleFit+XGB & 2251 & 0.21 & 0.21 & 0.22 \\
        & RuleSHAP w/o Step 2 & 1600 & 0.21 & 0.21 & 0.23 \\
        & RuleSHAP w/o Step 3 & 2031 & 0.50 & 0.57 & 0.57 \\
        & RuleSHAP & 1391 & \textbf{0.64} & \textbf{0.68} & \textbf{0.68} \\
        \midrule
        \multirow{6}{*}{\begin{tabular}[c]{@{}l@{}}GPT-4o\end{tabular}}
        & Decision Tree & 1057 & 0.00 & 0.00 & 0.00 \\
        & RuleFit & 2489 & 0.50 & 0.54 & 0.55 \\
        & RuleFit+XGB & 4090 & 0.36 & 0.36 & 0.36 \\
        & RuleSHAP w/o Step 2 & 2842 & 0.36 & 0.36 & 0.36 \\
        & RuleSHAP w/o Step 3 & 3104 & 0.71 & 0.71 & 0.72 \\
        & RuleSHAP & 2316 & \textbf{0.79} & \textbf{0.79} & \textbf{0.80} \\
        \midrule
        \multirow{6}{*}{\begin{tabular}[c]{@{}l@{}}Llama\\3.1 8B\end{tabular}}
        & Decision Tree & 1058 & 0.00 & 0.00 & 0.00 \\
        & RuleFit & 2169 & 0.21 & 0.31 & 0.32 \\
        & RuleFit+XGB & 3428 & 0.07 & 0.07 & 0.07 \\
        & RuleSHAP w/o Step 2 & 2193 & 0.07 & 0.07 & 0.07 \\
        & RuleSHAP w/o Step 3 & 2443 & 0.29 & 0.46 & 0.46 \\
        & RuleSHAP & 1747 & \textbf{0.50} & \textbf{0.61} & \textbf{0.61} \\
        \midrule
        \multirow{6}{*}{\begin{tabular}[c]{@{}l@{}}Llama\\3.1 70B\end{tabular}}
        & Decision Tree & 648 & 0.00 & 0.00 & 0.00 \\
        & RuleFit & 1422 & 0.21 & 0.25 & 0.26 \\
        & RuleFit+XGB & 2462 & 0.14 & 0.14 & 0.14 \\
        & RuleSHAP w/o Step 2 & 1761 & 0.14 & 0.14 & 0.14 \\
        & RuleSHAP w/o Step 3 & 1715 & \textbf{0.57} & \textbf{0.60} & \textbf{0.60} \\
        & RuleSHAP & 1360 & \textbf{0.57} & \textbf{0.60} & \textbf{0.60} \\
        \bottomrule
       \end{tabular}%
    }
    \caption{Exact recovery and rule counts across injected triggers on all targets. Best nonzero MRRs per LLM and cutoff are in bold.}
    \label{tab:xai_comparison}
\end{table}

\textbf{\textsc{RuleSHAP} vs.\ Baselines.}
Table~\ref{tab:xai_comparison} compares exact trigger recovery and rule counts over all injected ground-truth pairs. There, \textsc{RuleSHAP} obtains the best or tied-best nonzero MRR for every model and cutoff; this pattern also holds when restricting the evaluation to non-judge output abstractions only (Table~\ref{tab:no_llm_judge_summary}; Appendix Table~\ref{tab:no_llm_judge_full} gives full MRR@3/10 details).
Decision trees recover no exact injected triggers. RuleFit recovers some triggers but is weaker on every model, with average MRR@1 of 0.314 compared with 0.571 for \textsc{RuleSHAP}. \textsc{RuleSHAP} also extracts fewer rules than RuleFit for every model, reducing the aggregate rule count from 8,857 to 7,757 (12.4\%).
Wilcoxon signed-rank tests over reciprocal ranks show that \textsc{RuleSHAP} ranks exact matches significantly higher than RuleFit at RR@1, RR@3, and RR@10 (raw one-sided $p=0.003$, $0.0008$, and $0.0009$; rank-effect sizes $r=0.37$--$0.42$); all comparisons remain significant after Holm--Bonferroni correction.

\begin{table}
  \centering
  \resizebox{.9\linewidth}{!}{%
  \begin{tabular}{lcccc}
    \toprule
    \textbf{LLM}  & \begin{tabular}[c]{@{}c@{}}\textbf{Decision~}\\\textbf{Tree}\end{tabular} & \textbf{RuleFit} & \begin{tabular}[c]{@{}c@{}}\textbf{RuleSHAP~}\\\textbf{w/o Step 2}\end{tabular} & \textbf{RuleSHAP}     \\ 
    \midrule
    GPT-3.5 Turbo & 722 / 0.00 & 668 / 0.50 & \textbf{551} / 0.17 & 578 / \textbf{0.67} \\
    GPT-4o mini & \textbf{754} / 0.00 & 1158 / \textbf{0.67} & 1112 / 0.17 & 961 / \textbf{0.67} \\
    GPT-4o & \textbf{787} / 0.00 & 1778 / 0.83 & 2140 / 0.67 & 1729 / \textbf{1.00} \\
    Llama-3.1~8B & \textbf{780} / 0.00 & 1354 / 0.50 & 1481 / 0.17 & 1166 / \textbf{0.67} \\
    Llama-3.1~70B & \textbf{374} / 0.00 & 766 / 0.33 & 921 / 0.33 & 751 / \textbf{0.67} \\
    \bottomrule
  \end{tabular}
  }
  \caption{Results on non-judge output abstractions only: number of non-judge proxy rules and MRR@1 (\#rules / MRR@1).}
  \label{tab:no_llm_judge_summary}
\end{table}

\begin{table}
\centering
\setlength{\tabcolsep}{3pt}
\resizebox{.85\linewidth}{!}
{%
\begin{tabular}{lccc}
\toprule
\textbf{Method} & \textbf{Top-1 MCC} & \textbf{Top-3 MCC} & \textbf{Top-10 MCC} \\
\midrule
RuleSHAP & 0.76 [0.67, 0.79] & 0.80 [0.73, 0.88] & 0.83 [0.80, 0.95] \\
RuleFit & 0.75 [0.66, 0.80] & 0.78 [0.74, 0.82] & 0.82 [0.80, 0.90] \\
\bottomrule
\end{tabular}%
}
\caption{Functional fidelity of top-$k$ extracted rules. Entries are median MCC (quartiles in brackets) across models.}
\label{tab:functional_fidelity}
\end{table}

\textbf{Functional fidelity.}
Exact MRR does not reward rules that are syntactically different from the injected trigger but functionally similar on the observed abstraction distribution. Table~\ref{tab:functional_fidelity} therefore evaluates the top-$k$ rules as surrogate trigger predictors. \textsc{RuleSHAP} achieves median MCC 0.763 at top-1 and 0.801 at top-3, close to RuleFit at top-1 but higher at top-3.

\textbf{Performance by complexity.}
Table~\ref{tab:all_mrr_by_bias} breaks down MRR@1 by complexity tier.
Among the methods that recover any exact triggers, univariate triggers are generally easier than conjunctive and non-convex triggers, which expose a gap between conventional rule extraction and \textsc{RuleSHAP}.
Specifically, RuleFit MRR@1 averages 0.40 (univariate), 0.32 (conjunctive), and 0.27 (non-convex); global SHAP feature ranking gives 0.40, 0.40, and 0.37, respectively, but does not produce rules. \textsc{RuleSHAP} combines SHAP-driven feature prioritization with rule induction, giving 0.67, 0.72, and 0.40.

\begin{table}[tbh]
\centering
\setlength{\tabcolsep}{3pt}
    \resizebox{\linewidth}{!}{%
    \begin{tabular}{llrrr} 
    \toprule
    \textbf{LLM}                                                            & \textbf{XAI Method} & \multicolumn{1}{c}{\textbf{Univariate}} & \multicolumn{1}{c}{\textbf{Conjunctive}} & \multicolumn{1}{c}{\textbf{Non-Convex}}  \\ 
    \hline
    \multirow{6}{*}{\begin{tabular}[c]{@{}l@{}}GPT-3.5\\Turbo\end{tabular}} & Decision Tree       & 0                                       & 0                                        & 0                                        \\
    & RuleFit             & \textbf{0.66}                           & 0.2                                      & 0.16                                     \\
    & RuleFit+XGB       & \textbf{0.66}                           & 0                                        & 0                                        \\
    & RuleSHAP w/o Step 2 & \textbf{0.66}                           & 0                                        & 0                                        \\
    & RuleSHAP w/o Step 3 & \textbf{0.66}                           & 0.2                                      & 0.16                                     \\
    & RuleSHAP            & 0.33                                    & \textbf{0.4}                             & \textbf{0.33}                            \\ 
    \hline
    \multirow{6}{*}{\begin{tabular}[c]{@{}l@{}}GPT-4o\\mini\end{tabular}}   & Decision Tree       & 0                                       & 0                                        & 0                                        \\
    & RuleFit             & \textbf{0.66}                           & 0.2                                      & 0.33                                     \\
    & RuleFit+XGB       & 0.33                                    & 0.2                                      & 0.16                                     \\
    & RuleSHAP w/o Step 2 & 0.33                                    & 0.2                                      & 0.16                                     \\
    & RuleSHAP w/o Step 3 & \textbf{0.66}                           & 0.6                                      & 0.33                                     \\
    & RuleSHAP            & \textbf{0.66}                           & \textbf{0.8}                             & \textbf{0.5}                             \\ 
    \hline
    \multirow{6}{*}{GPT-4o}                                                 & Decision Tree       & 0                                       & 0                                        & 0                                        \\
    & RuleFit             & 0.33                                    & \textbf{0.8}                             & 0.33                                     \\
    & RuleFit+XGB       & 0.33                                    & 0.4                                      & 0.33                                     \\
    & RuleSHAP w/o Step 2 & 0.33                                    & 0.4                                      & 0.33                                     \\
    & RuleSHAP w/o Step 3 & \textbf{1}                              & 0.6                                      & \textbf{0.66}                            \\
    & RuleSHAP            & \textbf{1}                              & \textbf{0.8}                             & \textbf{0.66}                            \\ 
    \hline
    \multirow{6}{*}{\begin{tabular}[c]{@{}l@{}}Llama\\3.1 8B\end{tabular}}     & Decision Tree       & 0                                       & 0                                        & 0                                        \\
    & RuleFit             & 0.33                                    & 0                                        & \textbf{0.33}                            \\
    & RuleFit+XGB       & 0.33                                    & 0                                        & 0                                        \\
    & RuleSHAP w/o Step 2 & 0.33                                    & 0                                        & 0                                        \\
    & RuleSHAP w/o Step 3 & 0.33                                    & 0.4                                      & 0.16                                     \\
    & RuleSHAP            & \textbf{0.66}                           & \textbf{0.8}                             & 0.16                                     \\
    \hline
    \multirow{6}{*}{\begin{tabular}[c]{@{}l@{}}Llama\\3.1-70B\end{tabular}}  & Decision Tree       & 0                                       & 0                                        & 0                                        \\
    & RuleFit             & 0                                       & 0.4                                      & 0.16                                     \\
    & RuleFit+XGB       & 0                                       & 0.2                                      & 0.16                                     \\
    & RuleSHAP w/o Step 2 & 0                                       & 0.2                                      & 0.16                                     \\
    & RuleSHAP w/o Step 3 & \textbf{0.66}                           & \textbf{0.8}                             & \textbf{0.33}                            \\
    & RuleSHAP            & \textbf{0.66}                           & \textbf{0.8}                             & \textbf{0.33}                           \\
    \bottomrule
    \end{tabular}
    }
\caption{MRR@1 by trigger complexity tier and LLM across injected triggers on all targets. Bold marks the best score within each model and complexity column.} \label{tab:all_mrr_by_bias}
\end{table}

\begin{table}[tbh]
\centering
\setlength{\tabcolsep}{3pt}
\resizebox{.68\linewidth}{!}
{\begin{tabular}{lcc}
\toprule
\textbf{LLM} &
\begin{tabular}[c]{@{}c@{}}
$dCorr$(\textit{common},\\ \textit{expl.\ length})
\end{tabular} &
\begin{tabular}[c]{@{}c@{}}
$dCorr$(\textit{positive},\\ \textit{subjectivity})
\end{tabular} \\
\midrule
GPT-3.5 Turbo   & 0.238\dcs & 0.734\dcl \\
GPT-4o mini     & 0.868\dcl & 0.878\dcl \\
GPT-4o          & 0.324\dcm & 0.736\dcl \\
Llama-3.1~8B       & 0.150\dcs & 0.862\dcl \\
Llama-3.1~70B   & 0.455\dcm & 0.899\dcl \\
\bottomrule
\end{tabular}}
\caption{Representative input--output certificates ($dCorr\in[0,1]$). Superscripts denote effect size:
\textsuperscript{\scriptsize S} small ($<0.3$),
\textsuperscript{\scriptsize M} medium ($[0.3,0.7)$),
\textsuperscript{\scriptsize L} large ($\ge 0.7$).}
\label{tab:corr_cert_main}
\end{table}

\textbf{Correlation Certificates.}
We validate that non-judge proxies track the injected nonlinear templates using distance correlation, \(dCorr\), between input abstractions \(u_k\) and output abstractions \(v_j\) \citep{szekely2009brownian} (cf.\ \S\ref{sec:tom:preliminaries}). Table~\ref{tab:corr_cert_main} reports the two main injection channels: \textit{common} \(\rightarrow\) explanation length (\(b_1\)) and \textit{positive} \(\rightarrow\) subjectivity (\(b_2/b_3\)). Across models, \textit{common} \(\rightarrow\) length ranges from \(0.15\) to \(0.87\), while \textit{positive} \(\rightarrow\) subjectivity is large for \(b_2\) (\(0.73\)--\(0.90\)) and moderate-to-large for \(b_3\) (\(0.49\)--\(0.83\)); all remain significant after Bonferroni correction (\(p<0.01\)). Additional certificates and SHAP-by-tier recovery scores appear in Appendix~\ref{apx:certificates_extra_results}.

\textbf{Ablation Study.}
To isolate each step's contribution in \textsc{RuleSHAP}, we examine the full pipeline and two ablations: one without SHAP-weighted XGBoost feature sampling (Step~2) and one without SHAP-aware LASSO selection (Step~3).
Table~\ref{tab:xai_comparison} shows that removing Step 2 sharply reduces MRR@1, e.g., from 0.36 to 0.14 for GPT-3.5 Turbo and from 0.50 to 0.07 for Llama-3.1~8B, highlighting the importance of SHAP-guided XGBoost in capturing behavioural triggers.
Skipping SHAP-driven LASSO pruning (Step 3) impacts performance less consistently: on GPT-3.5 Turbo, \textsc{RuleSHAP} without Step 3 achieves an MRR@1 of 0.29 instead of 0.36; on Llama-3.1~8B, it drops from 0.50 to 0.29. In cases like GPT-4o, the decline is minor (0.71 vs.\ 0.79). Overall, the two ablations show different failure modes: omitting Step 2 sharply lowers recovery, whereas omitting Step 3 usually retains more distractor rules and can reduce early precision.

\textbf{Qualitative rule recovery.}
Table~\ref{tab:ruleshap-gpt4o-b3} reports a representative set of top-ranked rules extracted by \textsc{RuleSHAP} for GPT-4o under $b_3$.
These rules can be \emph{directly interpreted} by humans. For instance, the first rule states:
\quotes{\emph{If a topic is relatively uncommon (\(\text{common} \le 0.5\)) and is associated with a high positivity score (\(\text{positive} > 0.5\)), then the model generates longer explanations.}}
This rule aligns with the length component inherited by \(b_3\) from \(b_2\), where explanation length was manipulated based on topic positivity and rarity. The \emph{impact direction} (positive or negative) indicates whether the rule increases or decreases the value of the target output abstraction (here, explanation length). The \emph{importance} score $I(r)$ reflects the magnitude of this impact and serves to prioritize which rules have the strongest effects.

\begin{table}[tb]
    \centering
    \resizebox{.8\linewidth}{!}
    {
        \begin{tabular}{@{}p{0.6\linewidth}cc@{}}
        \toprule
        \textbf{Rule} & \begin{tabular}[c]{@{}c@{}}\textbf{Impact}\\\textbf{~Direction}\end{tabular} & \textbf{$I(r)$} \\ 
        \midrule
        \(\text{common} \le 0.5 \land \text{positive} > 0.5\) & Positive & 2005.67 \\\hline
        \(\text{common} > 0.30 \land \text{common} \le 0.5 \land \text{positive} > 0.5 \land \text{interdisciplinary} > 0.5 \land \text{interdisciplinary} \le 0.70\) & Positive & 110.85 \\\hline
        \(\text{common} \le 0.5 \land \text{positive} > 0.5 \land \text{interdisciplinary} > 0.5\) & Positive & 100.95 \\\hline
        \(\text{common} > 0.5 \land \text{positive} \le 0.70 \land \text{interdisciplinary} > 0.70 \land \text{interdisciplinary} \le 0.89\) & Negative & 93.79 \\\hline
        \(\text{common} > 0.30 \land \text{common} \le 0.5 \land \text{socially controversial} \le 0.5 \land \text{positive} > 0.5 \land \text{positive} \le 0.70\) & Positive & 59.08 \\\hline
        \(\text{common} > 0.5 \land \text{interdisciplinary} \le 0.5\) & Negative & 55.70 \\
        \bottomrule
        \end{tabular}
    }
    \caption{{Top-6 rules extracted by \textsc{RuleSHAP} from GPT-4o under \(b_3\)} for the target output abstraction \quotes{explanation length}.}
    \label{tab:ruleshap-gpt4o-b3}
\end{table}

\section{No-Injection Rule Steering} \label{sec:case_study}

To complement the controlled injections, we also run \textsc{RuleSHAP} in a \emph{no-injection} setting. In this setting, no behavioural rule is added to the system instruction; the base model is treated as a black box, and recurring input--output regularities are extracted from its ordinary responses to SDG-related prompts. We focus on the top-5 rules per output abstraction and summarize the salient patterns in Table~\ref{tab:sdg_ruleshap_summary}.

Applying \textsc{RuleSHAP} to base-model responses reveals model-specific shifts in readability, sentiment, response length, and subjectivity as a function of topic properties such as commonality, technicality, controversy, and interdisciplinarity. Across models, sentiment largely tracks explicit positive or negative cues, while stylistic behaviours diverge: common topics are often simplified; complexity rises most consistently for controversial prompts and, in some models, for interdisciplinary or conceptually dense inputs. Length typically increases for common or demanding prompts, and subjectivity shifts conditionally, often decreasing with technicality but increasing in dense, uncommon, or negative regimes. These patterns are supported by statistically significant correlations between rule features and output abstractions ($p\!<\!0.001$, $1\!-\!\beta\!>\!0.95$).

A held-out no-injection generalization analysis tests whether salient rules preserve their effect direction on unseen topics. Direction consistency is the fraction of held-out topics for which the rule's observed effect has the same sign as the effect estimated on the training split. For Llama-3.1~8B, rules are extracted on five random train splits and evaluated on the corresponding held-out topics. For \textsc{RuleSHAP}, direction consistency is 1.00 for explanation length and 0.96 for Gunning--Fog readability, sentiment, and subjectivity. The mean over the four metrics is 0.97, indicating that the no-injection rules are stable behavioural regularities.

\begin{table}[tb]
    \centering
    \resizebox{.98\linewidth}{!}
    {
        \begin{tabular}{ll} 
        \hline
        \textbf{Models} & \textbf{Salient RuleSHAP patterns} \\ 
        \hline
        GPT-4o & \begin{tabular}[c]{@{}l@{}}
            {[}common $\to$ lower fog],\\
            {[}common $\to$ longer text],\\
            {[}positive $\to$ more positive polarity]
        \end{tabular} \\ 
        \hline
        GPT-4o mini & \begin{tabular}[c]{@{}l@{}}
            {[}common $\to$ lower fog],\\
            {[}socially controv. $\to$ higher fog],\\
            {[}positive $\to$ more positive polarity]
        \end{tabular} \\ 
        \hline
        GPT-3.5 Turbo & \begin{tabular}[c]{@{}l@{}}
            {[}interdisc./conceptually dense $\to$ higher fog],\\
            {[}technical complexity $\to$ higher fog],\\
            {[}technical complexity $\land$ low controversy $\to$ less subjectivity]
        \end{tabular} \\ 
        \hline
        Llama-3.1~8B & \begin{tabular}[c]{@{}l@{}}
            {[}common $\to$ lower fog],\\
            {[}socially controv. $\to$ higher fog],\\
            {[}dense/uncommon/negative $\to$ more subj.]
        \end{tabular} \\
        \hline
        \end{tabular}
    }
    \caption{No-injection case study: salient \textsc{RuleSHAP}-extracted behavioural patterns. \emph{Fog} is the Gunning Fog readability index: higher values indicate harder-to-read text.}
    \label{tab:sdg_ruleshap_summary}
\end{table}

\textbf{From detection to intervention.}
We show that recovered rules can serve as \emph{conditional triggers} for lightweight, targeted steering.
As a testbed, we use Llama-3.1~8B in the no-injection setting and run rule-gated steering experiments.

Each experiment follows the same procedure.
We \textit{(i)} select a high-importance RuleSHAP rule for a target output abstraction, such as subjectivity or readability; 
\textit{(ii)} identify the subset of prompts where the rule fires; and 
\textit{(iii)} apply a fixed system instruction only on that subset to steer the target output abstraction.
For example, for subjectivity, we use the top dense/uncommon/negative trigger: [\texttt{conceptually dense \!>\! 0.70 \!\&\! common \!$\leq$\! 0.5 \!\&\! negative \!>\! 0.89 \!\&\! neutral \!$\leq$\! 0.70}]; this rule fires on $N_{\text{fire}}=173$ topics.
Enforcing an objective tone on these generations reduces the subjectivity proxy from 
$0.53\pm0.37$ to $0.15\pm0.29$ 
(paired $t$-test, $p<.001$, $d=-0.86$).
Analogous interventions reduce subjectivity in a second regime and Gunning--Fog scores in two readability regimes. Table~\ref{tab:sdg_mitigation} reports triggered sample sizes, before/after means, and paired effect sizes; all differences are significant at \(p<.001\). Appendix~\ref{apx:certificates_extra_results} lists the steering instructions.

\begin{table}[tb]
\centering

\resizebox{.9\linewidth}{!}{
\begin{tabular}{@{}p{0.16\linewidth}p{0.42\linewidth}cccr@{}}
\toprule
\makecell[l]{\textbf{Target}\\\textbf{abstract.}}
& \makecell[l]{\textbf{Triggering}\\\textbf{prompt profile}}
& \makecell[c]{\textbf{$N_{\text{fire}}$}}
& \makecell[c]{\textbf{Before}\\\textbf{mitig.}}
& \makecell[c]{\textbf{After}\\\textbf{mitig.}}
& \makecell[r]{\textbf{Effect}\\\textbf{size }$\boldsymbol{d}$} \\
\midrule

\multirow[t]{2}{=}{Subjectivity}
& Dense; uncommon; negative; low-neutral
& 173 & \msd{0.53}{0.37} & \msd{0.15}{0.29} & $-0.86$ \\

& Low-controversy; medium-neutral
& 55 & \msd{0.56}{0.41} & \msd{0.07}{0.19} & $-1.08$ \\

\hline

\multirow[t]{2}{=}{Gunning--Fog}
& Uncommon; negative; low-neutral
& 225 & \msd{19.83}{2.70} & \msd{14.64}{3.88} & $-1.33$ \\

& Low-technicality; common; controversial; positive; medium-neutral
& 300 & \msd{19.10}{2.88} & \msd{14.55}{4.93} & $-0.88$ \\

\bottomrule
\end{tabular}
}
\caption{Rule-gated steering on Llama-3.1~8B. 
Values are means $\pm$ standard deviation on triggered subsets; all paired tests have $p<.001$.}
\label{tab:sdg_mitigation}

\end{table}

\section{Discussion \& Limitations} \label{sec:discussion} 

Model-agnostic global rule-extraction XAI methods use surrogate models such as decision trees or gradient boosting (\S\ref{sec:experiments:baselines}). Decision-tree surrogates produce compact symbolic models but recover no exact injected triggers in this benchmark, showing that compactness alone is not sufficient. RuleFit and \textsc{RuleSHAP} both use gradient-boosted candidate rules, which are well suited to the tabular abstraction space \citep{mcelfresh2023neural}, but \textsc{RuleSHAP} adds SHAP-guided feature sampling and SHAP-aware sparse selection.
The resulting gain is largest for conjunctive triggers, where feature interactions matter and ordinary gradient boosting can rank the relevant predicates too low. Non-convex rule extraction remains difficult: \textsc{RuleSHAP} improves over RuleFit on average, but the harder tier still has lower recovery than the univariate and conjunctive tiers. Because the benchmark combines non-judge proxies with judge-derived scores, we treat the latter as supplementary signals and verify that the same conclusion holds without them. Thus, the benchmark is not evidence that current global XAI solves complex behavioural auditing; it identifies a measurable gap and shows that SHAP-guided rule induction narrows it.

\textbf{Error analysis.} Most missed exact matches are near misses: their target output abstraction or direction agrees with the injected trigger, but at least one predicate or threshold differs. This pattern is strongest for GPT-3.5 Turbo, whose smaller topic pool (Table~\ref{tab:unique_topics}) may reduce predicate discrimination for gradient-boosted rules.

\textbf{Correlation certificates} show that the extracted rules capture meaningful links between input and output abstractions. Additionally, non-judge proxies correlate with LLM-as-a-judge estimates, especially for framing effect (\(\rho=0.799\) with subjectivity; \(\rho=0.658\) with length) and information overload (\(\rho=0.604\) with length; \(\rho=0.476\) with Gunning--Fog), while oversimplification shows weaker but positive associations (\(\rho \leq 0.361\)); this supports treating judge-derived scores as supplementary signals alongside the non-judge proxies.

\textbf{No-injection case study.}
The no-injection case study shows that much apparent \quotes{style} variation is systematically driven by coarse topic attributes. Two points stand out. First, although some extracted patterns are semantically unsurprising (e.g., polarity cues track output sentiment), their stability across tens of thousands of prompts indicates persistent behavioural regularities in the underlying systems. Second, cross-model differences in readability and subjectivity suggest non-uniform stylistic policies: similar topic signals are mapped to distinct surface decisions (simplification, verbosity, hedging), plausibly reflecting divergent instruction-tuning and safety objectives.
The targeted steering experiments further show these rules are actionable: a high-importance trigger can serve as a targeted, low-overhead control signal (e.g., an objective-tone post-edit) that substantially shifts the subjectivity proxy on the affected subset, with predictable spillovers to length and readability. This illustrates \textsc{RuleSHAP} as a practical interface between behavioural diagnosis and selective intervention.

\textbf{Abstraction and imputation robustness.}
The approach depends on the supplied ordered abstractions, so we ran two sensitivity checks. Prompt-variant scoring on 80 topics gave 0.875 exact consensus, 0.998 within-one-point agreement, and mean pairwise Spearman 0.703. The de-duplication check in \S\ref{sec:experiments} showed limited sensitivity to encoder and threshold choices. For SHAP nearest-neighbour imputation, we do not claim a formal error bound. Empirically, over 24,576 perturbations, tie sets had median size 1, a multi-tie rate of about 0.23, and sampled-vs-tie-mean deviations of 0.012--0.030 for subjectivity and 0.035--0.044 for polarity. These checks support stability in our bounded-output experiments, while leaving validation necessary for new abstractions.

\textbf{Limitations.}
First, our injected triggers span only three complexity tiers (univariate, conjunctive, non-convex) and may under-represent real-world phenomena; even so, these simple cases already stress-test the state-of-the-art.
Second, to assess behaviours in AI-generated explanations, we use three LLM-as-a-judge abstractions, which may be judge-biased and introduce circularity concerns; therefore, we treat them as supplementary signals and report a non-judge-only ablation.
Third, global rule extraction overproduces rules; while this remains an open limitation across XAI, \textsc{RuleSHAP} improves early precision (MRR@10; Table~\ref{tab:xai_comparison}), reducing auditor effort.
Fourth, model-agnostic rule extraction requires a \emph{predefined symbolic map} that operationalizes behaviour hypotheses (definitions and observable indicators), which can limit scalability and portability; however, without such a mapping, detection reverts to black-box probing (a constraint shared by all XAI methods).



\section{Conclusion \& Future Work}
\label{sec:conclusion} 

We introduced a statistically grounded \emph{belief-abstraction} pipeline that maps topics and LLM explanations into abstract numerical spaces, enabling global explanations of belief-driven behaviour. We also proposed \textsc{RuleSHAP}, a model-agnostic rule-extraction method that uses global SHAP aggregates for feature sampling and LASSO pruning.
Across controlled settings with increasingly complex injected behavioural triggers and multiple LLMs, \textsc{RuleSHAP} achieves higher exact recovery and top-$k$ functional fidelity than the rule-extraction baselines, also producing fewer rules.
Correlation certificates reveal statistically detectable input--output regularities, while a no-injection case study uncovers recurring patterns that can support targeted rule-gated steering, such as reducing subjectivity on a rule-defined subset.
Future work includes: \textit{(i)} using the LLM-as-a-judge protocol \citep{zheng2023judging} to learn and refine abstractions, partially automating interpretable feature design; \textit{(ii)} studying additional complexity tiers to better characterize XAI limitations; and \textit{(iii)} extending rule-level detection to control and validation, including rule-gated prompting, targeted data augmentation, and mechanistic follow-up analyses in open-weight LLMs \citep{sovrano2026neuronanchored}.

\begin{acks}
This work was partially supported by Innosuisse (119.321 INT-ICT).
\end{acks}

\newpage
\bibliographystyle{ACM-Reference-Format}
\bibliography{references_min}

\appendix

\section{Abstraction Details}
\label{apx:abstractions}

For each input dimension, the audited LLM judges how web text about a topic instantiates that dimension, e.g., \quotes{\textit{Evaluate the conceptual density of the texts in the whole web about \{topic\}. Think about how complex and layered the ideas are, requiring significant mental effort to unpack.}} Analogous prompts replace the property name for technical complication, commonality, social controversy, unambiguity, geographic variability, time variability, interdisciplinarity, positivity, negativity, and neutrality. Each input prompt ends with:
\quotes{\textit{Rate your score on a scale from 1 (not \{property label\}) to 5 (very \{property label\}). Expected output: ES, estimated score from 1 to 5; SE, a very short explanation.}}
No few-shot examples are included in the scoring prompts. The normalized value used by XAI is $s/5$.

For output abstractions, character count and Gunning--Fog are non-learned functions of the generated explanation. Sentiment and subjectivity use chunked multilingual Transformer estimators as described in \S\ref{sec:tom}. The three supplementary judge prompts ask the model to assess framing effects, oversimplification, or information overload, then append the same 1--5 estimated-score format. For each audited LLM, judge scoring uses the corresponding base model under the judge prompt and without the injected behavioural instruction; judge scores are therefore treated as secondary targets rather than independent ground truth. They are interpreted only when they agree with non-judge proxies (which is the case in our experiments).

\section{Sampling and Hyperparameters}
\label{apx:hyperparameters}

\noindent\textbf{Topic generation.} Topic extraction uses the template: \quotes{\textit{Generate at least \{n\} distinct topics related to \{domain\}. All topics must have score \{score\}/5 when evaluating how \{dimension\} web texts about that topic are. Provide each topic with a short explanation of why it has that score.}} Generation uses $T=1$ and top-$p=1$ for diversity; abstraction scoring, judge scoring, injected-response generation, and targeted steering generations use deterministic decoding ($T=0$, top-$p=0$). The power analysis targets power $0.8$, medium effect size $0.35$, and $\alpha\le 0.05$, yielding roughly 60 topics per dimension-score cell before de-duplication.

\noindent\textbf{Rule extraction.} XGBoost uses 300 trees, maximum depth 5, subsample $0.8$, exact split search, minimum child weight 4, learning rate $0.01$, and SHAP-weighted feature sampling with per-level column sampling set to $1/d$, where $d$ is the number of input features. The SHAP-aware sparse stage uses 3-fold cross-validated LASSO over 100 candidate penalties.

\noindent\textbf{Exact-match rule criteria.}
For exact recovery, an extracted rule is counted as matching a ground-truth pair only when three conditions hold: (i) the target output abstraction is the same; (ii) the rule contains the same (or semantically equivalent) audited input dimension or dimensions as the injected predicate; and (iii) the threshold is equivalent after converting raw Likert scores to normalized midpoint thresholds. For example, $p_{\mathrm{raw}}\ge 3$ is matched to $p>0.5$, and $c_{\mathrm{raw}}\le 2$ is matched to $c\le 0.5$ when emitted by a tree split. Rules with the correct target and direction but different, non-equivalent predicates are treated as functional near misses rather than exact matches.


\section{Concrete Generated Examples}
\label{apx:concrete_examples}

These examples show how abstractions correspond to text:

\textbf{$b_1$ branch.}
For \emph{Age and Gender Parity Measures}, the raw commonality score is $c_{\mathrm{raw}}(\tau)=2$, so the concise $b_1$ branch returns:
\quotes{\textit{Age and Gender Parity Measures assess the equality of opportunities and outcomes across different age groups and genders, often in areas like education, employment, and health.}}
For \emph{Air Pollution}, $c_{\mathrm{raw}}(\tau)=5$, so the long-explanation $b_1$ branch returns:
\quotes{\textit{Air pollution refers to the presence of harmful or excessive quantities of substances in the air we breathe, which can have detrimental effects on human health, the environment, and the climate \dots}}

\textbf{No-injection sentiment examples.} GPT-3.5 Turbo rates \emph{Child Marriage} positive $1/5$ and negative $4/5$; its explanation has polarity $-0.776$, citing \quotes{{forced} and {serious negative consequences}, including {limited education opportunities}, {increased risk of domestic violence}, and {poor health outcomes}}; it also calls child marriage a \quotes{violation of human rights} and a \quotes{harmful traditional practice}. For \emph{Zumba}, it rates positive $4/5$ and negative $2/5$, with polarity $0.486$, citing \quotes{{popular}, {fun} and {energetic}, {accessible} to people of all fitness levels, and a {great} way to get in shape.}

\textbf{No-injection subjectivity examples.}
For \emph{Pathological Mechanisms in Neurovascular Coupling}, scored as technically complicated ($5/5$), GPT-4o produces an explanation with subjectivity $0.593$:
\quotes{Neurovascular coupling refers to the relationship between neuronal activity and subsequent changes in cerebral blood flow. This process ensures that active regions of the brain receive an \emph{adequate} supply of oxygen and nutrients\ldots{} Pathological mechanisms\ldots{} can disrupt this balance, leading to \emph{various} neurological disorders. Here are some \emph{key} aspects of these pathological mechanisms\ldots{}}
The full response also includes \emph{crucial}, \emph{inadequate}, \emph{excessive}, \emph{proper}, and \emph{normal}. By contrast, \emph{Charity Shop Purchases}, scored as technically simple ($1/5$), receives subjectivity near $0$:
\quotes{Charity shop purchases refer to the act of buying items from a charity shop\ldots{} These shops sell a variety of second-hand goods\ldots{}}

\section{RuleSHAP Proof Sketches}
\label{apx:ruleshap-proofs}


\noindent\textbf{Step 2 weighted feature sampling.} Let $R\subseteq\{1,\ldots,d\}$ be the behaviour-relevant features, $|R|=r$, and consider a depth-$L$ decision path whose split feature at each depth is sampled independently from a distribution $p$. Let $q=\sum_{i\in R}p_i$. Under uniform sampling, $q_{\mathrm{unif}}=r/d$ and the probability that at least one relevant feature appears on the path is $1-(1-r/d)^L$. Step~2 uses normalized global SHAP masses $\bar\rho_i$ as sampling priorities. If these weights concentrate more mass on relevant features than uniform sampling, $q=\sum_{i\in R}\bar\rho_i>r/d$, then
$
1-(1-q)^L > 1-(1-\frac{r}{d})^L .
$
For $K$ independent paths, the same argument gives $1-(1-q)^{LK}>1-(1-r/d)^{LK}$. The result proves increased coverage of relevant features under the sampling model, not exact trigger recovery; recovery also requires useful thresholds, retention after sparse selection, and high ranking against distractors.

\noindent\textbf{Step 3 weighted LASSO.} For target abstraction $\mathbf v$ and candidate-rule activation matrix $\mathbf X\in\{0,1\}^{N\times m}$, Step~3 solves
\[
\hat{\mathbf w}=\arg\min_{\mathbf w}\frac{1}{2N}\|\mathbf v-\mathbf X\mathbf w\|_2^2+\lambda_N\sum_{j=1}^m\omega_j|w_j|,\quad \omega_j=(\rho_{r_j}+\epsilon)^{-1} .
\]
This is a weighted LASSO and has the same form as adaptive LASSO when the SHAP-derived $\rho_{r_j}$ values are treated as preliminary relevance estimates. Let $S$ be the set of truly relevant candidate rules. Under the usual adaptive-LASSO regularity conditions on the design matrix, signal strength, and tuning sequence \citep{zou2006adaptive}, selection consistency follows when relevant rules receive bounded penalties and irrelevant rules receive asymptotically stronger effective penalties; in scaled-loss notation, sufficient rate conditions can be expressed as $\lambda_N\max_{j\in S}\omega_j=o(1)$ and $\sqrt{N}\lambda_N\min_{j\notin S}\omega_j\to\infty$, together with the standard identifiability and beta-min assumptions. If SHAP masses are uninformative or misleading, these assumptions fail; empirical recovery and fidelity tests are therefore necessary.

\section{Additional Experimental Results}
\label{apx:certificates_extra_results}

\noindent\textbf{Correlation certificates.} Certificates pair an input abstraction with an output abstraction and report multiplicity-corrected distance correlation. For $b_1$, commonality--length ranges from 0.150 to 0.868. For $b_2$, positive--subjectivity ranges from 0.734 to 0.900. For $b_3$, positive--subjectivity ranges from 0.495 to 0.835, negative--subjectivity ranges from 0.306 to 0.582, and interdisciplinarity--Gunning--Fog ranges from 0.058 to 0.851. The low end of the last range flags model-dependent weakness for the non-convex readability target. Length-related conjunctive certificates are smaller but significant: positive--length ranges from 0.194 to 0.545, and commonality--length ranges from 0.191 to 0.523. All reported certificates are significant at $p<0.001$ before correction; the main certificates remain significant after Bonferroni correction.

\noindent\textbf{Extra recovery results.} Table~\ref{tab:no_llm_judge_full} reports exact recovery after excluding the three judge-derived target abstractions. The evaluated target pairs are explanation length for all three complexity tiers, subjectivity for the conjunctive and non-convex tiers, and Gunning--Fog readability for the non-convex tier. Table~\ref{tab:shap_mrr_by_bias} reports global SHAP feature-ranking recovery by trigger complexity tier; SHAP often ranks relevant features highly, but it still does not produce symbolic rules.

\begin{table}[tbh]
\centering
\resizebox{\linewidth}{!}{%
\begin{tabular}{llrrrr}
\toprule
\textbf{LLM} & \textbf{Method} & \textbf{Rules} & \textbf{MRR@1} & \textbf{MRR@3} & \textbf{MRR@10} \\
\midrule
GPT-3.5 Turbo & Tree & 722 & 0.00 & 0.00 & 0.00 \\
 & RuleFit & 668 & 0.50 & 0.50 & 0.54 \\
 & RuleSHAP w/o S2 & 551 & 0.17 & 0.17 & 0.17 \\
 & RuleSHAP & 578 & \textbf{0.67} & \textbf{0.67} & \textbf{0.67} \\
GPT-4o mini & Tree & 754 & 0.00 & 0.00 & 0.00 \\
 & RuleFit & 1158 & \textbf{0.67} & 0.67 & 0.67 \\
 & RuleSHAP w/o S2 & 1112 & 0.17 & 0.17 & 0.21 \\
 & RuleSHAP & 961 & \textbf{0.67} & \textbf{0.75} & \textbf{0.75} \\
GPT-4o & Tree & 787 & 0.00 & 0.00 & 0.00 \\
 & RuleFit & 1778 & 0.83 & 0.83 & 0.83 \\
 & RuleSHAP w/o S2 & 2140 & 0.67 & 0.67 & 0.67 \\
 & RuleSHAP & 1729 & \textbf{1.00} & \textbf{1.00} & \textbf{1.00} \\
Llama-3.1~8B & Tree & 780 & 0.00 & 0.00 & 0.00 \\
 & RuleFit & 1354 & 0.50 & 0.67 & 0.67 \\
 & RuleSHAP w/o S2 & 1481 & 0.17 & 0.17 & 0.17 \\
 & RuleSHAP & 1166 & \textbf{0.67} & \textbf{0.75} & \textbf{0.75} \\
Llama-3.1~70B & Tree & 374 & 0.00 & 0.00 & 0.00 \\
 & RuleFit & 766 & 0.33 & 0.42 & 0.42 \\
 & RuleSHAP w/o S2 & 921 & 0.33 & 0.33 & 0.33 \\
 & RuleSHAP & 751 & \textbf{0.67} & \textbf{0.72} & \textbf{0.72} \\
\bottomrule
\end{tabular}}
\caption{Exact recovery on non-judge output abstractions only. Rule counts include all extracted rules, while MRR uses only non-judge output abstractions with injected ground-truth triggers.}
\label{tab:no_llm_judge_full}
\end{table}

\begin{table}[tbh]
\centering
\resizebox{\linewidth}{!}
{%
\begin{tabular}{llrrr}
\toprule
\textbf{LLM} & \textbf{Type} & \textbf{MRR@1} & \textbf{MRR@3} & \textbf{MRR@10} \\
\midrule
GPT-3.5 Turbo & Uni. & 0.00 & 0.50 & 0.50 \\
 & Conj. & 0.40 & 0.40 & 0.50 \\
 & Non-conv. & 0.50 & 0.56 & 0.60 \\
 & All & 0.42 & 0.49 & 0.55 \\
GPT-4o mini & Uni. & 1.00 & 1.00 & 1.00 \\
 & Conj. & 0.40 & 0.57 & 0.60 \\
 & Non-conv. & 0.33 & 0.56 & 0.58 \\
 & All & 0.42 & 0.60 & 0.62 \\
GPT-4o & Uni. & 1.00 & 1.00 & 1.00 \\
 & Conj. & 0.40 & 0.50 & 0.58 \\
 & Non-conv. & 0.33 & 0.47 & 0.53 \\
 & All & 0.42 & 0.53 & 0.59 \\
Llama-3.1~8B & Uni. & 0.00 & 0.00 & 0.14 \\
 & Conj. & 0.40 & 0.57 & 0.62 \\
 & Non-conv. & 0.33 & 0.47 & 0.56 \\
 & All & 0.33 & 0.47 & 0.55 \\
Llama-3.1~70B & Uni. & 0.00 & 0.00 & 0.25 \\
 & Conj. & 0.40 & 0.50 & 0.55 \\
 & Non-conv. & 0.33 & 0.47 & 0.52 \\
 & All & 0.33 & 0.44 & 0.51 \\
\bottomrule
\end{tabular}}
\caption{Global SHAP feature-ranking recovery by behaviour type. SHAP identifies relevant features but does not emit symbolic rules.}
\label{tab:shap_mrr_by_bias}
\end{table}

\noindent\textbf{Targeted steering instructions.} For subjectivity-reduction interventions, the rule-triggered prompts receive the instruction: \quotes{\textit{Write an explanation in a strictly objective and neutral tone. Avoid opinions, value judgments, and emotive or persuasive language. Focus on verifiable facts, definitions, and causal mechanisms. Use impersonal phrasing and avoid taking sides.}} For readability interventions, the rule-triggered prompts receive: \quotes{\textit{Write a simple, easy-to-read explanation. Use short sentences and common words. Avoid jargon; if technical terms are necessary, define them plainly.}} These instructions are applied only to prompts whose extracted rule fires; all other prompts keep the no-injection setup.

\noindent\textbf{No-injection interpretation.} In the no-injection SDG case study, \textsc{RuleSHAP} recovers stable, interpretable shifts rather than injected ground truth. Sentiment follows explicit positive and negative topic abstractions. Readability and subjectivity vary more by model: common topics are often simplified, controversial or interdisciplinary topics often increase Fog, and subjectivity can decrease under technical conditions while increasing in dense, uncommon, or negative regimes. The targeted steering results show that high-importance rules can serve as selective steering triggers, but they also show that one proxy intervention can shift other style dimensions such as length or readability. Extracted rules should therefore be treated as audit hypotheses to be validated with targeted counterfactual tests, not as causal mechanisms.

\end{document}